\documentclass[conference]{IEEEtran}

\usepackage{graphicx} 
\usepackage{subfigure}
\usepackage{amsthm}
\usepackage{amsmath} 
\usepackage{amssymb}  
\usepackage{color}
\usepackage{multirow}
\usepackage{balance}
\usepackage{cite}
\usepackage{algpseudocode}
\usepackage{algorithm}
\usepackage{algorithmicx}
\usepackage{url}

\begin{document}
%
\title{Power Systems Data Fusion based on \\ Belief Propagation}

\author{\IEEEauthorblockN{Francesco Fusco, Seshu Tirupathi and Robert Gormally}
\IEEEauthorblockA{IBM Research Ireland\\
Dublin, Ireland\\
\{francfus,seshutir,robertgo\}@ie.ibm.com}}


\maketitle

\begin{abstract}
The increasing complexity of the power grid, due to higher penetration of distributed
resources and the growing availability of interconnected, distributed metering devices 
requires novel tools for providing a unified and consistent view of the system.
A computational framework for power systems data fusion, based on probabilistic graphical models, 
capable of combining heterogeneous data sources with classical state estimation nodes and other customised computational nodes, 
is proposed. The framework allows flexible extension of the notion of grid state 
beyond the view of flows and injection in bus-branch models, and an efficient, naturally distributed inference algorithm can be derived. 
An application of the data fusion model to the quantification of distributed solar energy is proposed through 
numerical examples based on semi-synthetic simulations of the standard IEEE 14-bus test case.
\end{abstract}

\IEEEpeerreviewmaketitle

\section{Introduction}\label{sec:introduction}

The electrical grid is going through a significant transformation
towards a more distributed architecture for demand-supply balancing, due to
a higher penetration of distributed sources of renewable generation, storage
and demand flexibility. Internet-of-Things (IOT) technologies are an integral
part of the transformation, with energy utilities availing of more and more
highly-distributed intelligent devices which produce an ever-increasing amount of
heterogeneous data significantly different in terms of
format, resolution and quality \cite{Yu2015,Collier2017}. 

Effective management and coordination of such increased complexity requires scalable 
ingestion and fusion of all available data sources, from traditional
supervisory control and data acquisition (SCADA) systems to advanced metering infrastructure (AMI)
and machine learning models based on high-resolution weather data and forecasts \cite{Lonij2016,Fusco2016}.
In order to make best use of all the available information and support
robust advanced analytics and effective data-driven decision making, one of the key
challenges is to obtain a unified, consistent view of the electrical grid system
as a whole from the limited views, often overlapping but at times inconsistent,  
offered by the aforementioned set of data sources.

Traditionally, power systems state estimation is the tool designed to provide 
such a unified view of the grid for specific automation and control purposes \cite{Abur2004,Monticelli2000}. 
Extended state estimation has been proposed in the past to deal with the identification of unknown 
parameters of the bus-branch model \cite{Monticelli2000}. Various studies also proposed 
the idea of combining different data sources, such as smart meters or machine learning models, with the 
objective of providing regularization or pseudo-measurements to overcome unobservability \cite{Manitsas2008,Wu2012}. 
More recently, computational methods bades on probabilistic graphical models and belief propagation 
were studied to offer a naturally distributed solution of state estimation \cite{Hu2011,Cosovic2016,Cosovic2016a}. 

In this context, the state variable was designed to represent uniquely the set of voltages, 
flows and injections in a branch-bus model of the grid. 
Such view is, however, becoming obsolete as it does not provide sufficient observability into the system 
for both operation and planning purposes. As an example, system operators are more and more in need 
to quantify the impact of distributed resources, such as renewable energy or demand response, 
at a certain feeder or substation. Compiling such information may require combination of data from AMI and SCADA,
and necessitates of ad-hoc heuristics and domain expertise to resolve eventual gaps and inconsistencies between the 
data sources. 
A framework for power systems data fusion, based on probabilistic graphical models, is therefore proposed, where 
the set of state variables can be conveniently augmented as needed and a unified estimate based on all 
available data sources is obtained. 
A set of diverse computational nodes for each data source, including but not limited to classical state estimation algorithms,  
can be efficiently combined in a plug and play fashion, so that existing software modules
can be reused and the system can be extended in a modular way. 

After a description of the problem, in section \ref{sec:problem}, the computational framework is detailed in section \ref{sec:methods}. 
The main technical contribution of the paper, compared to existing methods proposed in \cite{Hu2011,Cosovic2016,Cosovic2016a}, 
is the derivation of a more general Gaussian belief propagation algorithm supporting multivariate, non-linear nodes, as described in section \ref{sec:methods_inference} 
Some numerical examples are proposed in section \ref{sec:results}.
\section{Problem Statement} \label{sec:problem}

Consider a set of data sources, $y_i \in \mathbb{R}^{m_i}$, with $i=1,\ldots\, M$, which may be related to 
different aspects of the power grid, described by a set of state variables $x_i \in \mathbb{R}^{n_i}$, with $i=1,\ldots\, M$,
as follows: 
\begin{equation}
 y_i = f_i(x_i)+ \varepsilon_i \quad i=1,\, 2,\ldots\, M.
\label{eq:data_sources}
\end{equation} 
In (\ref{eq:data_sources}), $f_i(\cdot)$ are generally non-linear, possibly known, relations between the data
and the state variables, while $\varepsilon_i$ represents the uncertainty in such relations or the noise in the data.

In power systems state estimation, a relation of the type in (\ref{eq:data_sources}) is used to relate the data
available from SCADA to a state variable, usually defined as the set of voltage magnitudes and phase angles at
the network buses in a bus-branch model, through power flow equations \cite{Monticelli2000,Abur2004}.

Recent trends have seen the growth of distributed, interconnected sensor devices, such as AMI. 
As a result, metering data of electrical consumption and distributed energy resources at 
individual premises, including generation but also controllable devices such as thermostats \cite{Callaway2011}, 
are increasingly made available. Extensions to the traditional notion of power system state, 
in the sense of set of physical quantities fully characterizing the behaviour of the system, 
beyond the abstraction of bus injections, can therefore be considered. 
For example, the fundamental contributions of solar generation or electric storage to a bus injection or
the effects of demand response on the voltage at a bus, can also be of interest. 
A relation of the type in (\ref{eq:data_sources}), where $x_i$ represents
the distributed generation from solar systems or the energy due to controllable thermal loads, could therefore
be available. A functional relation $f_i(\cdot)$ can be a simple identity where such resources are directly metered, 
or more complex data-driven or physical models where necessary 
(see \cite{Callaway2011} for a model of distributed thermal loads).  

Yet another source of data, of the type in (\ref{eq:data_sources}), can come from 
machine learning models of energy resources (demand, generation, etc.) based on 
weather, calendar and behavioural data. The importance of statistical modelling has been 
widely recognized in the context of energy forecasting for market bidding and 
network operation \cite{weron2007modeling}. Latest trends have seen the
need for obtaining forecasts at much finer spatial resolution due to an increased penetration of 
distributed resources \cite{Lonij2016}. Statistical load models have also been proposed in the context of load profiling 
or pseudo-measurement generation for unobservable state estimation problems \cite{Manitsas2008,Wu2012}. 

The individual functions of state estimation, AMI data processing and machine learning
can all benefit from each other in order to provide a more accurate, holistic view of the system, summarized by 
an appropriately defined set of state variables. 
By finding a mathematical relation of the type $g(x) = 0$, where $ x = \left\{x_1,x_2,\ldots\, x_M\right\}$, 
relating the state variables \emph{seen} by the different data sources, one could formalise the problem 
as solution of: 
\begin{align}
y &= f(x) + \varepsilon \label{eq:measurements_global}\\
0 &= g(x) + \varepsilon_0 \label{eq:constraints_global}.
\end{align} 
The problem in (\ref{eq:measurements_global}) and (\ref{eq:constraints_global}) will be 
referred to as the data fusion problem. Least-squares solutions, 
based on a Gauss-Newton iterative scheme, could be used. However, scalability can be an issue when 
the number of state variables or data sources grow. In addition, such a monolithic approach is not flexible, 
and the solution algorithm to the data fusion problem needs to be changed whenever a new data source or 
state variable is added or removed. By exploiting the natural factorization of the problem, 
a more flexible approach based on factor graphs and belief propagation is proposed in section \ref{sec:methods}.

\section{Methodology} \label{sec:methods}

A natural way to combine heterogeneous set of models like the ones described in section \ref{sec:problem} 
is provided by factor graphs. As a quite general class of graphical models, factor graphs provide a way to link 
potentially very different models together in a principled fashion that respects the roles of probability theory. 
An inference and learning algorithm can be obtained by linking together the modules and associated algorithms, in a plug and play fashion \cite{Frey2005}.
A brief introduction of factor graphs is provided in section \ref{sec:methods_intro} (see \cite{Loelinger2004,Loelinger2007,Barber2012} for more details). 
A general inference algorithm based on belief propagation is then derived in section \ref{sec:methods_inference}.

\subsection{Factor Graphs} \label{sec:methods_intro}

Given a set of variables, $\mathcal{Z}=\left\{z_1,z_2,\ldots\,,z_p\right\}$, factor graphs are a mathematical tool 
which can be used to express factorizations of the form: 
\begin{equation}
\Phi(z_1,z_2,\ldots\,,z_p) = \prod_{j=1}^q \phi_j(\mathcal{Z}_j),
\label{eq:factor_graph}
\end{equation}
where $\phi_j(\mathcal{Z}_j)$ is a factor defined on a subset of the variables $\mathcal{Z}_j \in \mathcal{Z}$. 
Graphically, (\ref{eq:factor_graph}) is represented by a node for each variable $z_i$ (a \emph{variable node}), 
and by a node for each factor $\phi_j$ (a \emph{factor node}). 
An edge between a factor node and a variable node exists for each $z_i \in \mathcal{Z}_j$. 

In the particular setting described in section \ref{sec:problem}, the set of variables $\mathcal{Z}=\left\{\mathcal{X},\mathcal{Y}\right\}$ 
can be specified by a set of $n$ random state variables $\mathcal{X} = \left\{x_1,x_2,\ldots\,,x_n\right\}$ and by a set of 
$m$ independent sources of noisy observations $\mathcal{Y} = \left\{y_1,y_2,\ldots\,,y_{m}\right\}$. 
The joint probability distribution over all the variables can then be factorized as: 
\begin{equation}
p(\mathcal{X},\mathcal{Y}) = \prod_{i=1}^m p(y_i|\mathcal{X}_{i}),
\label{eq:density_factorization}
\end{equation}
where $p(\cdot)$ denotes a probability density function, $p(y|\mathcal{X})$ denotes conditional probability and 
$\mathcal{X}_i \in \mathcal{X}$. With reference to the data fusion model defined in section \ref{sec:problem}, 
the measurement models in (\ref{eq:measurements_global}) 
define the conditional densities $p(y_i|\mathcal{X}_i)$, while the constraints in (\ref{eq:constraints_global})
can also be represented as conditional probabilities where $y_i=0$.

The objective of the data fusion model is to obtain an estimate of the posterior distribution of the 
unknown state variables with respect to a realization of the measurement data, which means solving the following inference problem: 
\begin{equation}
p(\mathcal{X}|\mathcal{Y}) = \dfrac{p(\mathcal{Y}|\mathcal{X})p(\mathcal{X})}{p(\mathcal{Y})}.
\label{eq:inference_problem}
\end{equation}

\subsection{Inference Algorithm based on Belief Propagation} \label{sec:methods_inference}
Belief propagation, or the sum-product algorithm, reduces the inference problem in (\ref{eq:inference_problem}) to 
the computation of localised messages between factor and variable nodes, 
along each edge of the factor graph defined by the factorization of the density 
function in (\ref{eq:density_factorization}). 

The messages are real-valued functions expressing the influence between variables. 
Messages from variables to factors are given by \cite{Barber2012}: 
\begin{equation}
\mu_{x_i\to \phi_j}(x_i) = \prod_{\phi_{k} \in \left\{\text{ne}(x_i) \setminus \phi_j\right\}} 
	\mu_{\phi_{k}\to x_i}(x_i),
\label{eq:message_to_factor}
\end{equation} 
where $\text{ne}(x)$ denotes the set of factors connected to the variable $x$. 
Note that in (\ref{eq:message_to_factor}) the dependency on $y$ was dropped since it will be treated 
as a known quantity, given that it represents evidence (measurement data), as described in section \ref{sec:methods_intro}. 
Messages from factors to variables are given by \cite{Barber2012}: 
\begin{equation}
\mu_{\phi_j \to x_i}(x_i) = \int_{w \in \left\{\mathcal{X}_j \setminus x_i\right\}}
	\phi_j(\mathcal{X}_j) \prod_{x_k \in \left\{\text{ne}(\phi_j)\setminus x_i\right\}} \mu_{x_k \to \phi_j(x_j)}.
\label{eq:message_to_variable}
\end{equation}
Once all the messages are updated, variables marginalization is computed as \cite{Barber2012}:
\begin{equation}
p(x_i) = \prod_{\phi_{k} \in \text{ne}(x_i)} \mu_{\phi_{k}\to x_i}(x_i).
\label{eq:marginal}
\end{equation}
Based on (\ref{eq:message_to_factor}), (\ref{eq:message_to_variable}) and (\ref{eq:marginal}), the inference
 is reduced to a sum (integral) of products of simpler terms than the ones in the full joint distribution, 
therefore the name sum-product algorithm. 

Computing the integral in (\ref{eq:message_to_variable}) is not feasible, in general, for continuous distribution and 
it can become numerically intractable for discrete distributions. 
By further assuming that the measurement factions and the state variables in (\ref{eq:data_sources}) 
represent Gaussian densities, the factorization in (\ref{eq:density_factorization}) can be expressed as follows: 
\begin{equation}
p(\mathcal{X},\mathcal{Y}) \propto \prod_{i=1}^m e^{-\frac{1}{2}\left[y_i-f_i(\mathcal{X}_i)\right]^\top R_i\left[y_i-f_i(\mathcal{X}_i)\right]},
\label{eq:density_factorization_gaussian}
\end{equation}
where $\propto$ denotes proportionality and $R_i$ is the covariance matrix of the error term $\varepsilon_i$ in (\ref{eq:data_sources}). 

Under the factorization in (\ref{eq:density_factorization_gaussian}), the messages in (\ref{eq:message_to_factor}) and (\ref{eq:message_to_variable})
are also Gaussian distributions defined, net of a normalizing constant, by a mean vector and a covariance matrix. 
The sum-product algorithm is then reduced to a set of simple linear algebra calculations.  

In particular, the algorithm can be derived by exploiting the canonical form of a Gaussian distribution 
and linearizing around an operational point:
\begin{equation}
e^{-\frac{1}{2}\left[y_i-f_i(\mathcal{X}_i)\right]^\top R_i\left[y_i-f_i(\mathcal{X}_i)\right]}
 \propto e^{-\frac{1}{2}\delta \mathcal{X}_i^\top J_i \delta \mathcal{X}_i +\delta \mathcal{X}_i^\top h_i },
 \label{eq:canonical}
\end{equation}
where $J_i = F_i^\top R_i^{-1} F_i$, with $F_i$ being the Jacobian of $f_i(\mathcal{X}_i)$ 
with respect to a point $\mathcal{X}_i=\overline{\mathcal{X}}_i$,
and $h_i = F_i^\top R_i^{-1} (y_i-f_i(\overline{\mathcal{X}}_i))$. Note that, in (\ref{eq:canonical}), 
the dependency from the constant term due to $y_i-f_i(\mathcal{X}_i)$ has been dropped. 

At a given iteration, $t$, assuming an estimate $x^t$ is available, then it is possible to compute the $J_j$ and $h_j$ for every factor, 
based on (\ref{eq:canonical}). Messages from variable $x_i$ to factor $f_j$ can then be computed as:
\begin{align}
h_{x_i\to f_j} &= \sum_{k \in \mathcal{K}_i\setminus j} h_{f_k\to x_i} \label{eq:msg2f_mu}\\
J_{x_i\to f_j} &= \sum_{k \in \mathcal{K}_i\setminus j}J_{f_k\to x_i}, \label{eq:msg2f_J}
\end{align}
where $\mathcal{K}_i$ is the set of factors $f_i$ connected to the variable $x_i$. 
Messages from factor $j$ to variable $i$ are, on the other hand, calculated based on: 
\begin{align}
h_{f_j\to x_i} &= h_{j} - \sum_{k \in \mathcal{K}_j\setminus i} J_{j}^{jk}(J_{x_k\to f_j} + J_j^{kk})^{-1}(h_{x_k\to f_j}+h_j^{k}) \label{eq:msg2v_mu}\\
J_{f_j\to x_i} &= J_{j} - \sum_{k \in \mathcal{K}_j\setminus i} J_{j}^{jk}(J_{x_k\to f_j} + J_j^{kk})^{-1}J_j^{kj}, \label{eq:msg2v_J}
\end{align}
where $J_j^{jk}$ is the block of the $J_j$ matrix with rows corresponding to the variable $x_j$ and columns 
corresponding to variable $x_k$. Similarly, $h_j^k$ denotes the block of the $h_j$ vector corresponding to the variable $x_j$. 

All messages (\ref{eq:msg2f_mu}) to (\ref{eq:msg2v_J}) can be computed by starting from the leaf factor nodes and
iteratively updating all messages where the required input messages have been processed. 
Once messages from all incoming factors are available, variable marginals can be updated with the following iterative scheme: 
\begin{align}
x_i^{t+1} &= x_i^t + \delta x_i \\
\delta x_i &= \left(\sum_{k \in \mathcal{K}_i} J_{f_k \to x_i}\right)^{-1} \left(\sum_{k \in \mathcal{K}_i} h_{f_k \to x_i}\right),
\label{eq:marginal_update}
\end{align}
with the corresponding covariance matrix given by:
\begin{equation}
S_{x_i}^{t+1} = \left(\sum_{k \in \mathcal{K}_i} J_{f_k \to x_i}\right)^{-1}.
\end{equation}
If the graph is not a tree, a loopy version of the proposed belief propagation algorithm can be derived
to iteratively converge to a solution for the marginals \cite{Cosovic2016}. For a full derivation of the fundamental Gaussian belief propagation messages 
in the linear case, the reader can refer to \cite{Loelinger2004,Loelinger2007}. The proposed algorithm is a 
generalization to non-linear factors and it is actually equivalent, in the case of trees, 
to a distributed implementation of the Gauss-Newton algorithm. 

In the context of power systems, the belief propagation algorithm proposed in \cite{Cosovic2016,Hu2011}
was limited to the solution of state estimation problems. To the best of the authors knowledge, 
a belief propagation algorithm for non-linear AC power flow models was only 
studied in \cite{Cosovic2016a}. The latter, however, involved a fully distributed algorithm 
on scalar factor nodes, resulting in heavily loopy graphs and requiring unclear heuristic rules to converge even for simple problems. 
The algorithm derived here can regarded as a generalization of the mentioned studies
where complete state estimation problems (or parts of) represent individual factor nodes, 
thus reducing the number of possible loops even for very large and complex grid topologies.
Additional non-linear measurement functions of any desired state variable can also be 
integrated into the model in a plug and play fashion. 
\section{Results}\label{sec:results}

\subsection{Available Data}
\begin{figure}
\centering
\includegraphics[width=7cm]{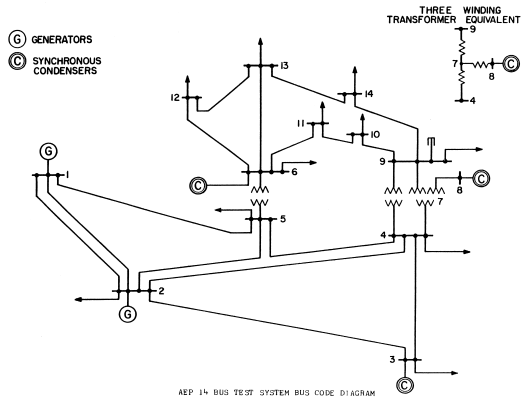}
\caption{The IEEE 14 Bus Test Case \cite{IEEEArchive}.}
\label{fig:14bus}
\end{figure}

\begin{figure}
\centering
\includegraphics[width=7cm]{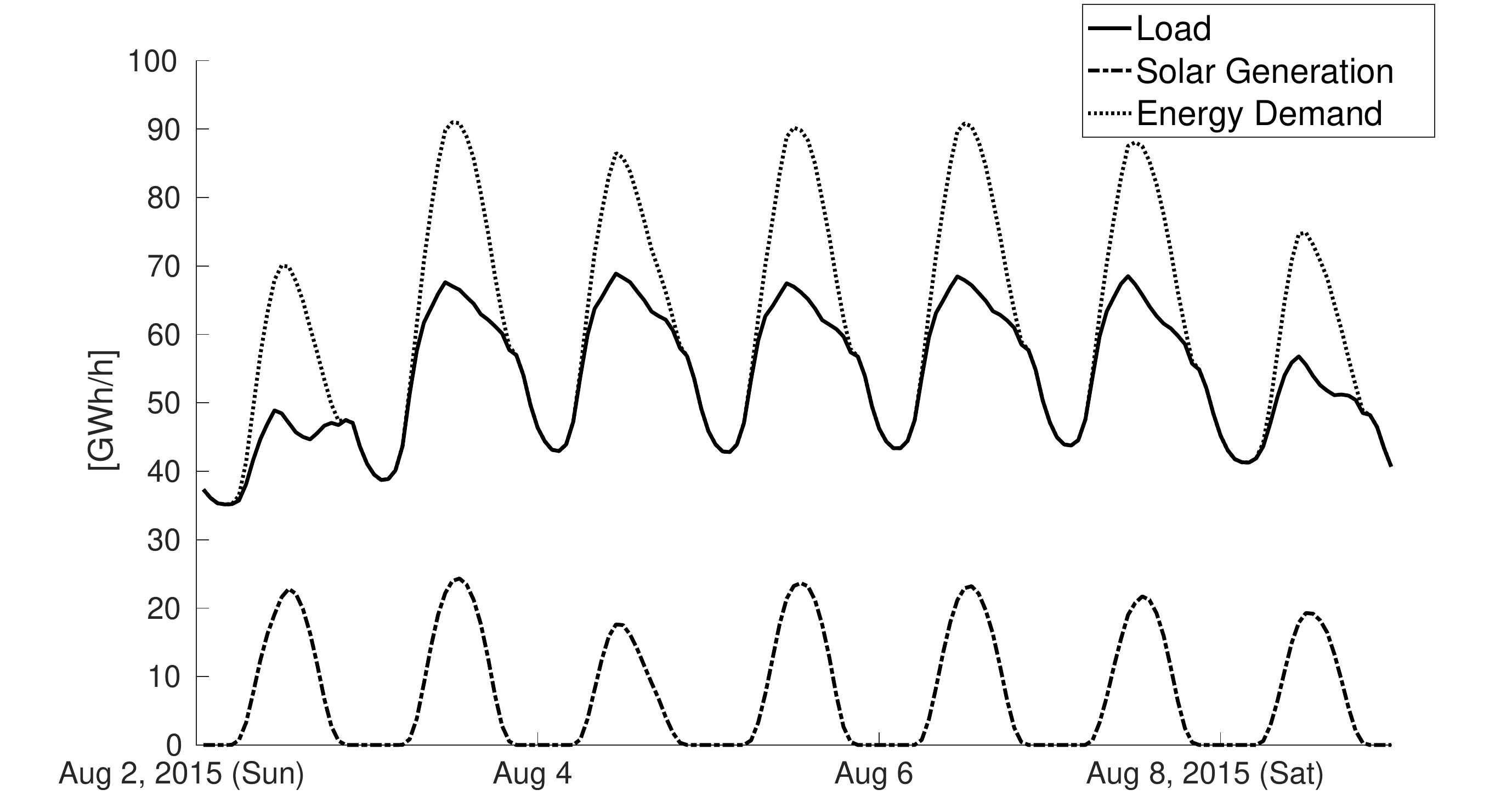}
\caption{Hourly Germany load and solar generation data from Aug 2, 2015 to Aug 8, 2015 \cite{OpenData2017}.}
\label{fig:data}
\end{figure}

Experiments were designed based on the standard 14-bus IEEE test case, shown in Fig \ref{fig:14bus}. 
The grid was simulated with time series data obtained from the Open Power Systems Data Platform \cite{OpenData2017}. 
In particular, national load and solar generation data from Germany from Jan 1, 2014 to Aug 8, 2015, sampled at hourly resolution were considered. 
Figure \ref{fig:data} shows a snapshot of the last week of the data, and also a calculation of the electrical demand from 
the net electrical load and the solar photovoltaic generation data. The additional component of the residual load, 
due to distributed wind generation, was not considered, although significant in the case of Germany, 
since renewable energy modelling is not the focus of the study.

Weather data were used in order to design the machine learning models. 
The Weather Company's Cleaned Observations application programming interface (API) \cite{TWC2017} provides cleaned,
interpolated, historical observations of a wide variety of weather variables. 
Version two of the API was used to retrieve hourly historical observations 
of surface temperature, diffuse horizontal radiation, direct normal irradiance, and downward solar radiation at
52.52°N, 13.61°E (Berlin) from Jan 1, 2014, to Aug 8, 2015. A more rigorous choice would have been to 
use a combination of points appropriately distributed across Germany. 
A simple approach was however favoured as it did not impact the objectives 
of the proposed experiments. 

\subsection{Experimental Setup for the Data Fusion Model} \label{sec:results_setup}

Based on the load and solar generation profiles shown in Fig. \ref{fig:data}, 
a realistic simulation of the 14-bus system is provided as follows.
The active injection at each network bus provided by the standard model in \cite{IEEEArchive} 
is taken to represent the peak load 
over the validation week (Aug 2 to Aug 8, 2015). The hourly profile is then modulated 
according to the load profile. It is further assumed that the solar generation is uniformly 
spread across the load buses 3 to 6 and 9 to 14, 
such to obtain the demand and generation components of the active load. 

\begin{figure}
\centering
\includegraphics[width=7cm]{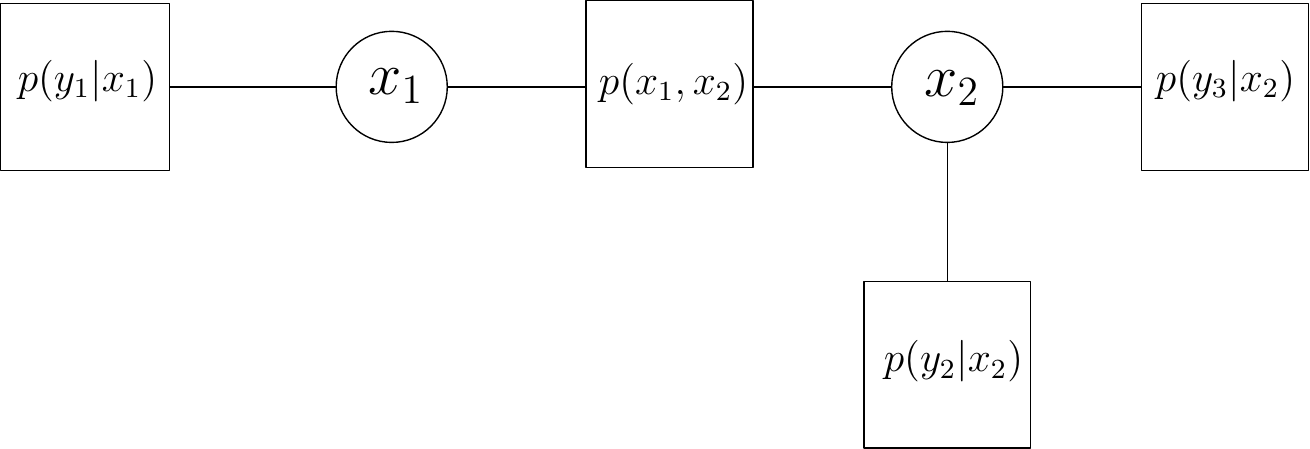}
\caption{Factor graph representation of the power systems data fusion model used in the experiments. 
Squares denote factors, circles are variable nodes. }
\label{fig:factor_graph}
\end{figure}

The data fusion model was then built based on the assumptions that data were generated 
from three different systems, producing the factor graph shown in Figure \ref{fig:factor_graph}.
Details of the factor nodes are outlined in the following.

\begin{figure}
\subfigure[]{\includegraphics[width=8.5cm]{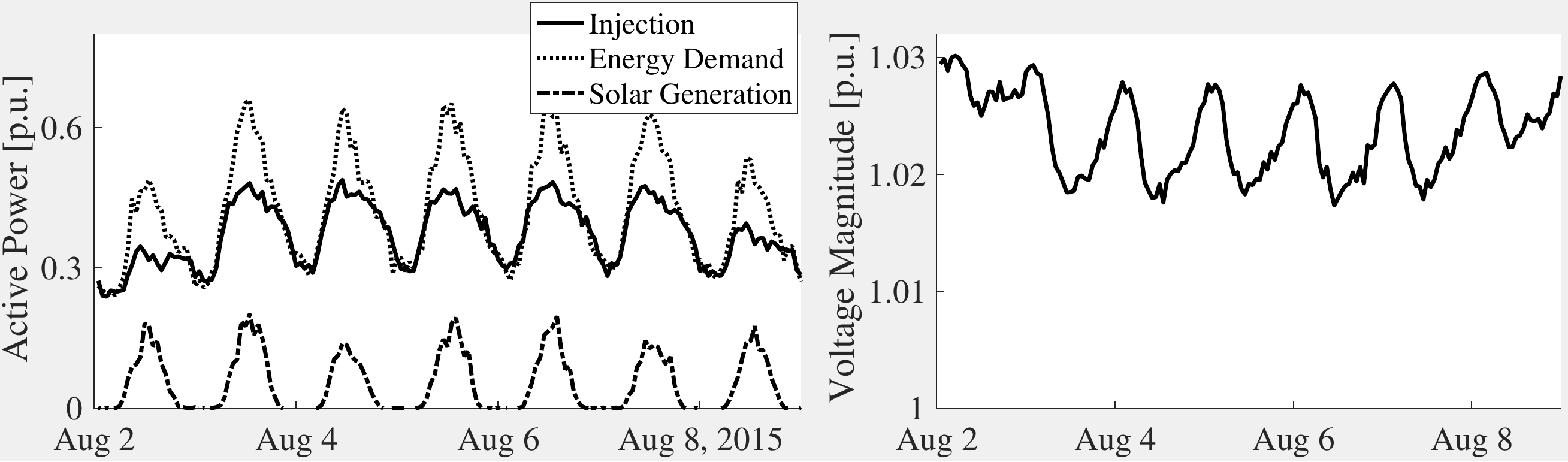}\label{fig:simulation1}}
\subfigure[]{\includegraphics[width=8.5cm]{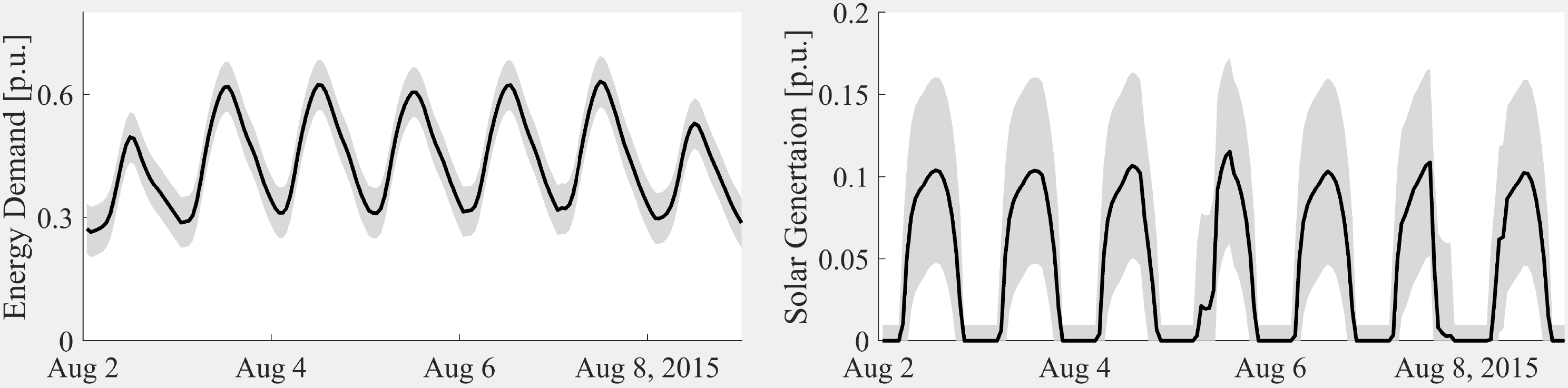}\label{fig:simulation2}}
\caption{Simulated data sources: \emph{(a)} Active power injection, energy demand, solar generation and
voltage magnitude at bus 4; \emph{(b)} Forecasts of energy demand and solar generation at bus 4, 
with the $95\%$ confidence interval. }
\vspace{-0.2cm}
\label{fig:simulation}
\end{figure}

\subsubsection{State Estimation Factor Node}
A set of data, $y_1$, is generated based on a power flow simulation 
of the 14-bus system. It is assumed that measurements are available for all active/reactive power 
injections and voltage magnitudes. Measurement data were generated by simulating the 14-bus power flow 
using Matpower \cite{Zimmerman2011} and the available load profiles. White Gaussian noise with standard deviation of 
$0.01$ p.u. for the power measurements and of $0.5e^{-3}$ p.u. for the voltage magnitudes was introduced. 
Figure \ref{fig:simulation1} shows an example of active power and voltage magnitude measurements generated at bus 4. 
The state variables, $x_1$, were defined as the set of bus voltages and measurement model, 
$y_1=f_1(x_1)+\varepsilon_1$, was based on 
the usual AC power flow equations \cite{Abur2004}. 

\subsubsection{Smart Meters Factor Node}
Smart meter data, $y_2$, were assumed to provide total energy demand and solar generation 
at the 10 load buses of the system, buses 3 to 6 and 9 to 14. 
White Gaussian noise with standard deviation of $0.02$ p.u. was introduced. Figure \ref{fig:simulation1} 
shows an example of the energy demand and generation data simulated at bus 4. 
The hidden state variable, $x_2$, was defined as representing the amount of energy demand and solar generation at each bus, 
so that the relation in (\ref{eq:data_sources}) is an identify matrix, namely $y_2=x_2 + \varepsilon_2$.  

\subsubsection{Machine Learning Factor Node}
Forecasting models providing data, $y_3$, of energy demand and solar generation at the 10 load buses specified in the 
state variable $x_2$ were also considered, so that the same model as for the smart meter factor node, 
$y_3=x_2 + \varepsilon_3$, was utilized.

The forecasts were computed using generalized additive models (GAMs) 
based on penalized iteratively re-weighted least squares method \cite{Ba2012,wood2015generalized}. 
Extensive research has gone into studying the covariates that are important to model electricity 
(see \cite{weron2007modeling,pierrot2011short,Ba2012} and 
references within). The energy demand model considered in this study is given by:
\begin{equation}
D =  s_1(\text{hour},\text{daytype} ) + s_2(T_m) + s_3(T_{max}) + \zeta \label{eq:load_gam} 
\end{equation}
where $D$ is the energy demand at a given time instant, $\text{daytype}$ is a categorical variable representing each day of a week,
$\text{hour}$ is the numerical variable representing the hour in a day, $T_m$ is the mean temperature for the 
 day and $T_{max}$ is the maximum temperature of the day. The basis functions $s_i(u)$
are smooth splines based on cubic B-splines. The solar generation forecast model was considered to be a linear model with normal and diffuse irradiance data as covariates. 

The models were trained on the period from January 1,
2014 to August 1, 2015, and a sample output for  bus 4 is shown in Fig. \ref{fig:simulation2}. 
A mean absolute percentage error for the demand forecasts of about $4.3\%$ was
observed on the validation week between August 2 and August 8, 2015. For the solar generation forecasts, 
a $10.4\%$ error was observed (normalized on the daily peak of generation to avoid numerical issues when generation is near the zero). 
Improved demand models could be designed by considering a more extensive set of features. 
Similarly, better solar generation forecasts can be obtained, 
for example by exploiting non-linead physics-based models \cite{Lonij2016}. 
However the focus of the paper is not on energy forecasting and such alternatives were not pursued. 

\subsubsection{Joint Factor Node}
A joint factor relating the state variables $x_1$ and $x_2$ was designed as: 
\begin{equation}
0 = f_{41} (x_1) - F_{42}x_2 + \varepsilon_{4},
\label{eq:joint_factor}\\
\end{equation}
where $f_{41}(x_1)$ represents the active injection equations of the AC power flow model, and
$F_{42}$ is a matrix where each row has a $+1$ and a $-1$ in correspondence, respectively, of the demand and generation entries in the state variable $x_2$
at the specific network bus. 
The Gaussian random error $\varepsilon_{4} \sim \mathbb{N}(0,R_{4})$ is chosen such thah $R_{4}$ is a diagonal matrix with very small diagonal entries of $1e-10$.

\subsection{Experimental results} \label{sec:results_experiments}

Since the factor graph contains no loops, as noted in section \ref{sec:methods_inference}, 
the proposed inference algorithm is equivalent to solving the global problem 
(\ref{eq:measurements_global})-(\ref{eq:constraints_global}) with the
Gauss-Newton method. However, instead of computing one matrix inversion of dimension $n_1+n_2$, 
four matrix inversions of dimension $n_1$ and $n_2$ are performed at every iteration, based on
(\ref{eq:msg2v_J}) and (\ref{eq:marginal_update}). 
As the number and dimensionality of the nodes increases, significant computational gain can be obtained. 

\begin{figure}
\centering 
\includegraphics[width=8.5cm]{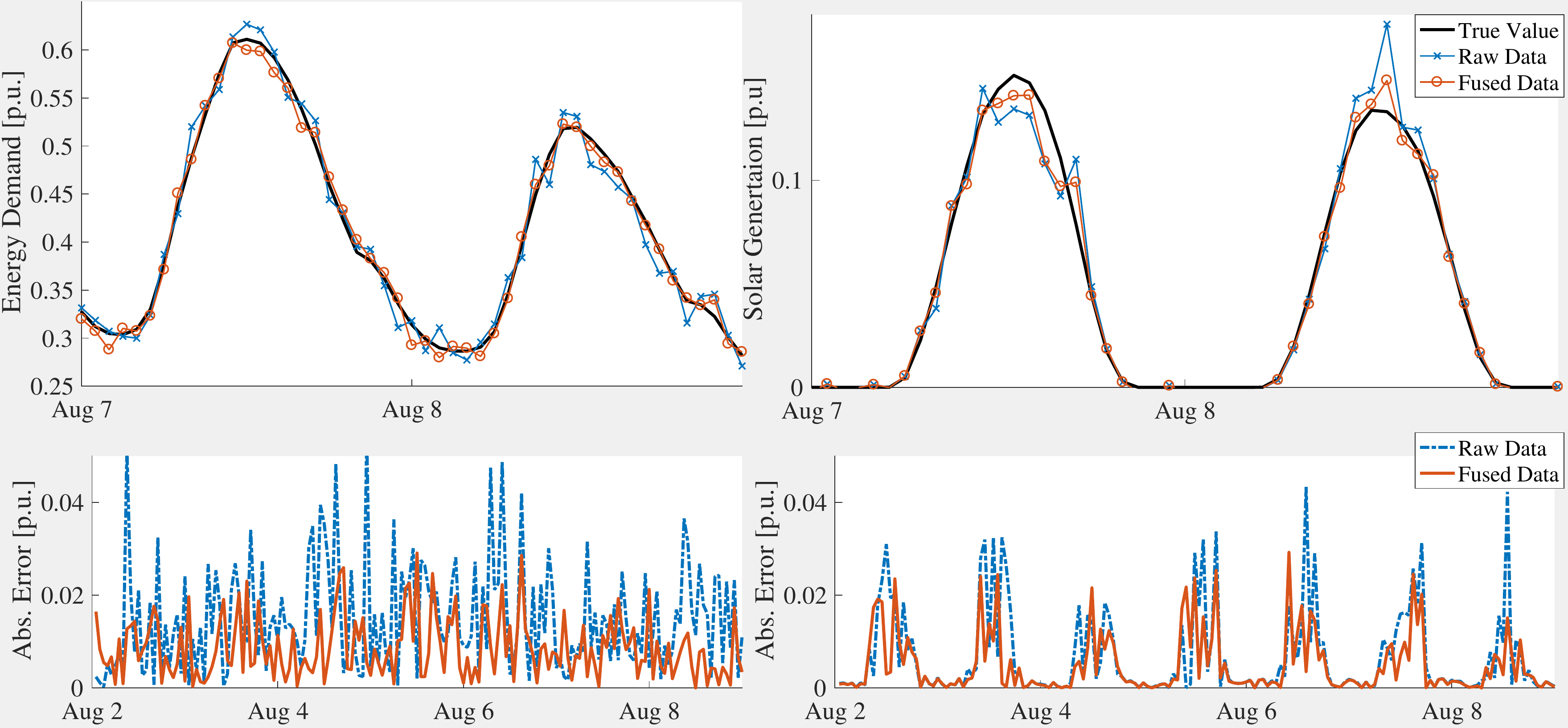}
\caption{Comparison between fused data and raw data of energy demand and solar genertaion at bus 4.}
\label{fig:exp1plot1}
\end{figure}

In a first experiment, the proposed inference algorithm was run on the simulated data 
over the validation week. Convergence was achieved in less then 10 steps for all 168 hourly samples. 
As expected, fusion of the three data sources allows reduction of uncertainty in estimating the true 
quantities with respect to using the information from the individual sources alone. 
As shown in Figure \ref{fig:exp1plot1}, the estimation of  
energy demand and solar generation at the bus 4, which is only metered directly from 
the smart meter data, also benefits from fusion with the state estimation node,
and shows lower error than when using the raw data only. Although not shown here, 
the improvement is consistent across all the network buses. 

\begin{figure}
\centering 
\includegraphics[width=7.5cm]{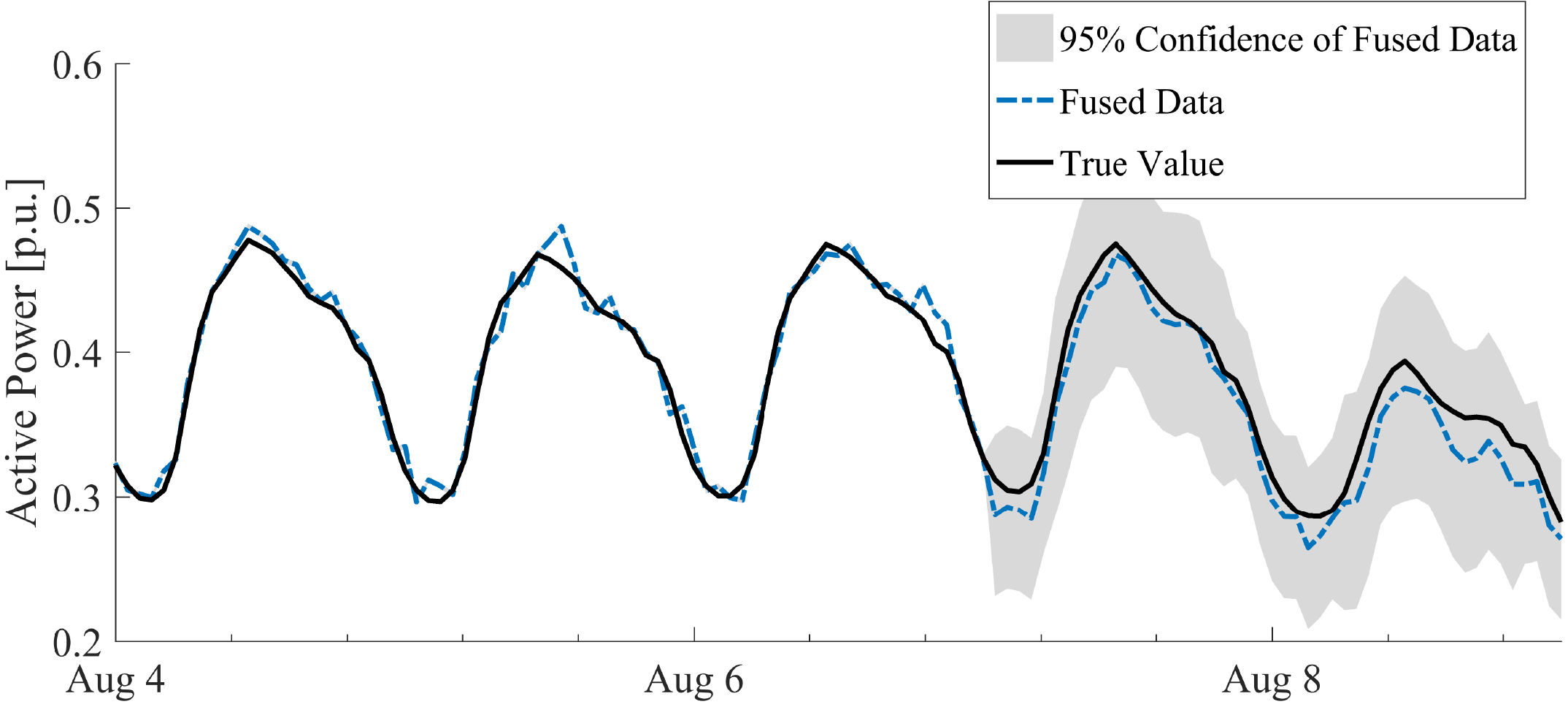}
\caption{Estimation of the active power injection at bus 4, with loss of observability in the state estimation node on August 7.
The confidence interval before Aug 7 is not visible as much smaller than after loss of observability.}
\label{fig:exp2plot1}
\end{figure}

In a second experiment, loss of observability on the last 2 days of the validation period
is considered in the state estimation node. In particular, voltages, active/reactive power and 
smart meter data are assumed unavailable from network buses 3, 4, 9 and 10 (bottom right of network diagram
in Fig. \ref{fig:14bus}). It can be shown that in this case the state estimation node
has a singularity, but the belief propagation algorithm does not experience invertibility issues since, 
through (\ref{eq:msg2f_J}) and (\ref{eq:msg2v_J}),
implicitly applies a regularization based on information available from the energy forecasts node.
Fused information is still obtained, although with higher uncertainty than when all data are available, 
as shown in Fig. \ref{fig:exp2plot1} for bus 4. 

\begin{figure}
\centering 
\includegraphics[width=7.5cm]{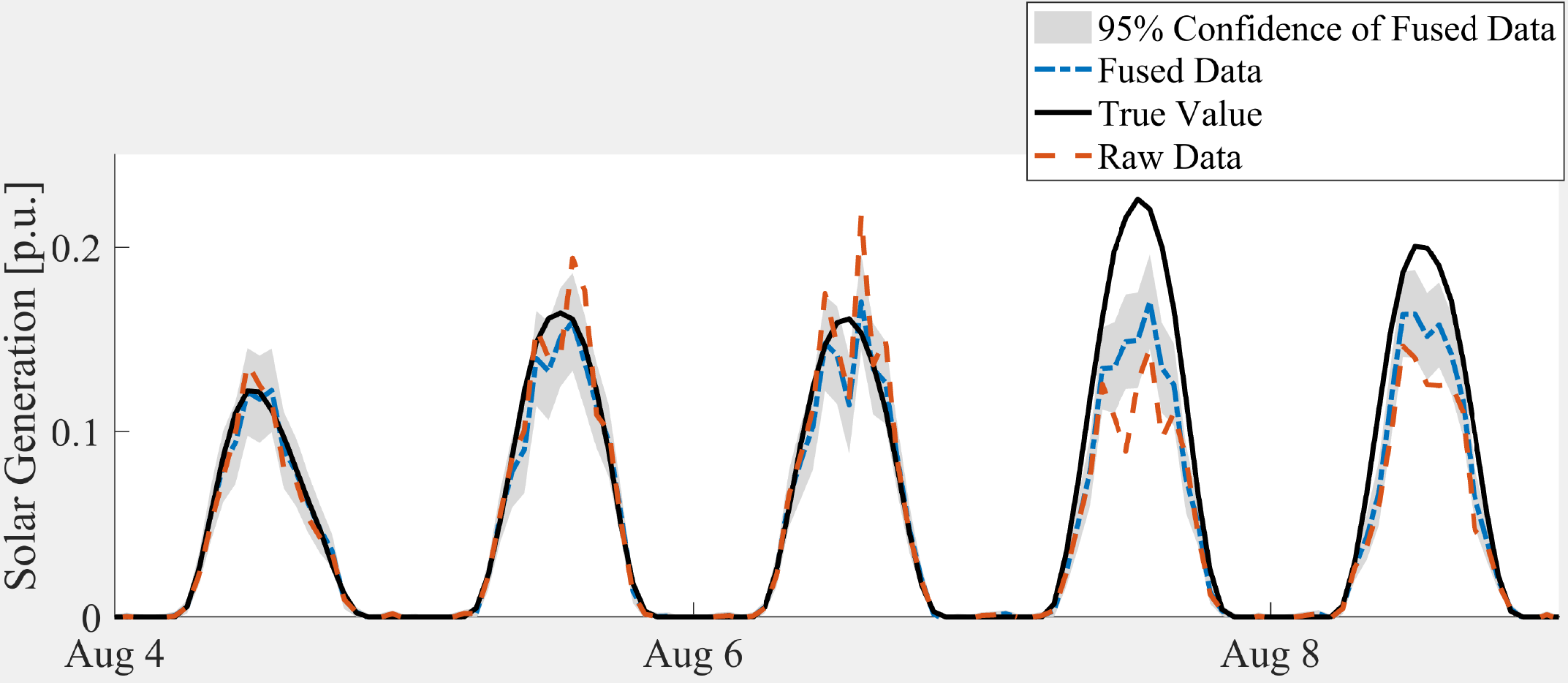}
\caption{Fused data and true value of solar generation at bus 4, when new solar capacity 
becomes available on August 7 but is not reflected in raw data. }
\label{fig:exp3plot1}
\vspace{-0.3cm}
\end{figure}

A final experiment shows how inconsistencies in a data source can be detected through data fusion. 
It is assumed that new solar installations cause an increase of $50\%$ capacity at bus 4 in the last 2 days 
of the validation period, and that this is not reflected in the smart meter data 
(actually, a typical real-world scenario). Figure \ref{fig:exp3plot1} shows how the fused estimate 
of solar generation at bus 4 is in larger than the raw data and closer to the true value by an amount 
comparable to the confidence interval. The possible inconsistency can be formally detected using statistical testing. 
Obviously such information comes from the grid data which are affected by the increased solar generation at the bus. 

\section{Conclusion} \label{sec:conclusion}

In this study, a novel computational framework for power systems data fusion was proposed. 
Based on probabilistic graphical models, heterogeneous data sources and related measurement models 
(physical laws or data-driven machine learning models) can be combined 
such that a consistent and unified view of the system is obtained. 
An efficient and naturally distributed inference algorithm based on Gaussian belief propagation is also derived.

In a set of numerical examples it was demonstrated how 
the traditional notion of state can be extended to provide visibility
into the amount of solar generation at the buses of a network model. 
The fused information reduces the effect of noise in the various data sources and
occurrences of missing or erroneous data are easily overcome and diagnosed. 

While fully known measurement functions were assumed, future work will investigate scenarios 
where they are partly unknown and need to be learned from the data. 
Further possibility for extensions of the state variable to 
include, for example, effects of temperature variation or of demand response programs on the active injections
will also be investigated. 
Since the data consistency is one the key benefits of data fusion, 
extensions of traditional bad data analysis to the proposed framework is also 
 of interest for future studies. 

\section*{Acknowledgments} \label{sec:acknowledgments}
This project has received funding from the European Research Council (ERC) under the European Union's Horizon 2020 research and innovation programme (grant agreement no. 731232).

\balance
\bibliographystyle{IEEEtran}
\bibliography{Energy}

\end{document}